\documentclass[preprint]{article}
\usepackage{spconf,amsmath,graphicx,hyperref}
\usepackage{booktabs}
\usepackage{multirow}
\usepackage{threeparttable}
\usepackage{array}
\usepackage{dblfloatfix} 
\usepackage[numbers,sort&compress]{natbib}
\usepackage{setspace}
\usepackage{amssymb}

\usepackage{tikz}

\makeatletter
\copyrightnotice{%
  \parbox[t]{\textwidth}{\footnotesize
  \textcopyright\ 2026 IEEE. Personal use of this material is permitted. 
  Permission from IEEE must be obtained for all other uses, in any current 
  or future media, including reprinting/republishing this material for 
  advertising or promotional purposes, creating new collective works, for 
  resale or redistribution to servers or lists, or reuse of any copyrighted 
  component of this work in other works.}%
}
\makeatother

\title{Cross-Representation Benchmarking in Time-Series Electronic Health Records for Clinical Outcome Prediction}
\name{Tianyi Chen, Mingcheng Zhu, Zhiyao Luo, Tingting Zhu}
\address{University of Oxford, United Kingdom}
\begin{document}
%
\maketitle

\begin{abstract}
Electronic Health Records (EHRs) enable deep learning for clinical predictions, but the optimal method for representing patient data remains unclear due to inconsistent evaluation practices. We present the first systematic benchmark to compare EHR representation methods, including multivariate time-series, event streams, and textual event streams for LLMs. This benchmark standardises data curation and evaluation across two distinct clinical settings: the MIMIC-IV dataset for ICU tasks (mortality, phenotyping) and the EHRSHOT dataset for longitudinal care (30-day readmission, 1-year pancreatic cancer). For each paradigm, we evaluate appropriate modelling families—including Transformers, MLP, LSTMs and Retain for time-series, CLMBR and count-based models for event streams, 8–20B LLMs for textual streams—and analyse the impact of feature pruning based on data missingness. Our experiments reveal that event stream models consistently deliver the strongest performance. Pre-trained models like CLMBR are highly sample-efficient in few-shot settings, though simpler count-based models can be competitive given sufficient data. Furthermore, we find that feature selection strategies must be adapted to the clinical setting: pruning sparse features improves ICU predictions, while retaining them is critical for longitudinal tasks. Our results, enabled by a unified and reproducible pipeline, provide practical guidance for selecting EHR representations based on the clinical context and data regime.
\end{abstract}

\section{Introduction}
\label{sec:intro}

\begin{figure}[!t] 
  \centering
  \includegraphics[width=\columnwidth]{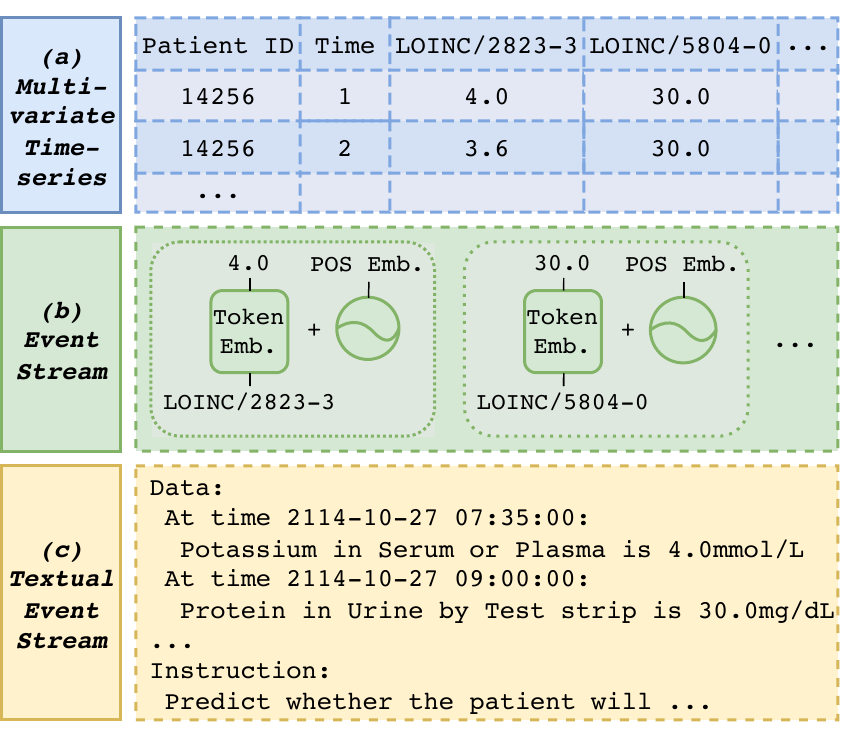}
  \caption{Different Representation Methods for EHRs. (a) multivariate time-series, with imputation; (b) event stream; (c) textual event stream, used for LLMs.}
  \label{fig:rep}
\end{figure}

Electronic Health Records (EHRs) capture diverse clinical measurements such as vital signs, laboratory results, medications, procedures, and diagnoses over time, making them a valuable resource for deep learning (DL) applications, such as survival analysis~\cite{zhang2024ai}, mortality prediction~\cite{wang2025artificial}, and rare disease prognosis~\cite{zhu2025bridging}.  However, fair evaluation remains challenging: performance depends not only on model choice but also on how the EHR is curated. Most prior work processed EHR as multivariate time-series data, which curates each patient onto fixed bins to form a matrix, and delivered influential intensive care unit (ICU) benchmarks, from multitask prediction on MIMIC to standardised feature engineering and imputation pipelines~\cite{harutyunyan2019multitask,gupta2022extensive}. But choices such as time window, feature aggregation, and missing value imputation can distort temporal information and introduce biases~\cite{xie2022deep}.

Recent research has pushed in two complementary directions. One transforms EHRs as event streams, an ordered sequence of timestamped clinical events. Models learn directly from irregular timing rather than from pre-aggregated bins; this view underlies CLMBR-style pretraining and broader event-centric modeling~\cite{arnrich2024medical,renc2024zero,wornow2023ehrshot}. The other transforms event streams into text by turning clinical events into descriptive sentences while preserving the temporal order~\cite{fleming2024medalign,zhu2025medtpe}. This textual event stream can be processed by large language models (LLMs) to make predictions. Examples of different representation methods for EHRS are shown in Fig.~\ref{fig:rep}.

Despite the emergence of new representation methods, there is no clear answer to which representation performs best. Direct comparison remains challenging due to inconsistent evaluation settings. For instance, Harutyunyan et al.\cite{harutyunyan2019multitask} evaluate MIMIC-III with 48-hour windows, Gupta et al.\cite{gupta2022extensive} use MIMIC-IV with different cohort definitions, while EHRSHOT~\cite{wornow2023ehrshot} focuses on non-ICU longitudinal tasks. Hence, the lack of controlled comparisons between representations, tasks, and clinical settings motivates us to design a benchmark that systematically evaluates multiple EHR representation methods using consistent data curation, evaluations, and identical splits across scenarios. We further examine how feature selection by missingness affects performance across representations.

The main contributions in this work are as follows: \\
(1) We present the first cross-representation benchmark of EHR data including multivariate time-series, event streams, and textual event streams, which enables evaluations for different representation methods. This benchmark supports two clinical prediction tasks per dataset: ICU mortality and ICU phenotyping for MIMIC-IV in the ICU setting, and 30-day readmission and 1-year pancreatic cancer for EHRSHOT in the longitudinal care setting. \\
(2) We build a reproducible unified pipeline that generates all three representations from raw EHR with cohorting and labelling for specific tasks, training, and evaluation under consistent splits and metrics. \\
(3) We provide key empirical insights. Models trained with the event stream outperform other representations overall; pretrained CLMBR is most sample-efficient in few-shot training, but simple count models can surpass it in all-shot settings with large sample size; missingness-based feature pruning removes many sparse variables with little performance loss, offering practical trade-offs between accuracy and simplicity.

\section{Methodology}

\begin{table*}[t]
\centering
\begin{threeparttable}
\small 
\setlength{\tabcolsep}{4pt} 
\renewcommand{\arraystretch}{0.85} 
\begin{tabular}{p{2cm}llcccccc} 
& & \multicolumn{3}{c}{\textbf{(a) MIMIC-IV}} & \multicolumn{3}{c}{} \\
\toprule
& & \multicolumn{3}{c}{\textbf{ICU Mortality}} & \multicolumn{3}{c}{\textbf{ICU Phenotyping}} \\
\textbf{Representation Method} & \textbf{Model} & \textbf{AUROC} & \textbf{AUPRC} & \textbf{F1} & \textbf{AUROC} & \textbf{AUPRC} & \textbf{F1} \\
\midrule
\multirow{4}{*}{\parbox{2cm}{\raggedright Multivariate Time-Series}} 
& Transformer & 0.806 & 0.264 & 0.244 & 0.700 & 0.370 & 0.401 \\
& MLP & 0.806 & 0.289 & 0.248 & 0.680 & 0.342 & 0.387 \\
& LSTM & 0.794 & 0.291 & 0.246 & 0.691 & 0.356 & 0.398 \\
& RETAIN & 0.790 & 0.302 & 0.251 & 0.686 & 0.349 & 0.392 \\
\midrule
\multirow{4}{*}{Event Stream} 
& Count (few-shot) & 0.530 & 0.074 & 0.112 & 0.553 & 0.248 & 0.250 \\
& CLMBR (few-shot) & 0.598 & 0.099 & 0.140 & 0.549 & 0.235 & 0.307 \\
& Count & 0.830 & 0.273 & 0.217 & \textbf{0.848} & \textbf{0.640} &  \textbf{0.600}\\
& CLMBR & \textbf{0.857} & \textbf{0.330} & 0.241 & 0.782 & 0.504 & 0.493 \\
\midrule
\multirow{4}{*}{\parbox{2cm}{\raggedright Textual Event Stream}} 
& GPT-OSS-20B & -- & -- & \textbf{0.254} & -- & -- & 0.256 \\
& Qwen3-8B-Thinking & -- & -- & 0.218 & -- & -- & 0.182 \\
& Llama3-8B & -- & -- & 0.137 & -- & -- & 0.184 \\
& DeepSeek-R1-8B & -- & -- & 0.135 & -- & -- & 0.218 \\
\midrule

& & \multicolumn{3}{c}{\textbf{(b) EHRSHOT}} & \multicolumn{3}{c}{} \\
\midrule
& & \multicolumn{3}{c}{\textbf{30-day Readmission}} & \multicolumn{3}{c}{\textbf{1-year Pancreatic Cancer}} \\
\textbf{Representation Method} & \textbf{Model} & \textbf{AUROC} & \textbf{AUPRC} & \textbf{F1} & \textbf{AUROC} & \textbf{AUPRC} & \textbf{F1} \\
\midrule
\multirow{4}{*}{\parbox{2cm}{\raggedright Multivariate Time-Series }} 
& Transformer & 0.666 & 0.222 & 0.296 & 0.610 & 0.055 & 0.049 \\
& MLP & 0.611 & 0.164 & 0.233 & 0.614 & 0.080 & 0.061 \\
& LSTM & 0.616 & 0.164 & 0.239 & 0.622 & 0.058 & 0.065 \\
& RETAIN & 0.617 & 0.162 & 0.238 & 0.636 & 0.061 & 0.069 \\
\midrule
\multirow{4}{*}{Event Stream} 
& Count (few-shot) & 0.679 & 0.208 & 0.266 & 0.674 & 0.077 & 0.074 \\
& CLMBR (few-shot) & \textbf{0.736} & \textbf{0.273} & \textbf{0.335} & 0.703 & 0.117 & 0.085 \\
& Count & 0.685 & 0.249 & 0.298 & \textbf{0.781} & \textbf{0.213} & \textbf{0.314} \\
& CLMBR& 0.691 & 0.225 & 0.278 & 0.706 & 0.103 & 0.163 \\
\midrule
\multirow{4}{*}{\parbox{2cm}{\raggedright Textual Event Stream}}
& GPT-OSS-20B & -- & -- & 0.232 & -- & -- & 0.179 \\
& Qwen3-8B-Thinking & -- & -- & 0.226 & -- & -- & 0.114 \\
& Llama3-8B & -- & -- & 0.209 & -- & -- & 0.045 \\
& DeepSeek-R1-8B & -- & -- & 0.211 & -- & -- & 0.128\\
\bottomrule
\end{tabular}
\caption{Performance comparison across three EHR data representations on MIMIC-IV and EHRSHOT datasets, where the F1 of LLM is calculated from 1000 bootstraps; the metrics of the ICU Phenotyping task are macro-averaged across 25 classes; the few-shot models are trained on 16 samples; bold font marks the best result per column.}\label{tab:main_results}
\end{threeparttable}
\end{table*}

\subsection{Datasets and Tasks}\label{sec:task_def}
We study two EHR corpora and two prediction tasks per corpus.
(1) \textbf{MIMIC-IV}~\cite{mimiciv22}  with \emph{ICU mortality} (binary) which predicts whether a patient will die during an ICU stay; and \emph{ICU phenotyping} (25 multi-labels), that predicts which acute care conditions are present from a patient's ICU admission to hospital discharge with phenotypes defined in~\cite{harutyunyan2019multitask}.
(2) \textbf{EHRSHOT}~\cite{wornow2023ehrshot} with binary classification tasks \emph{30-day readmission} and \emph{new diagnosis of pancreatic cancer} as defined in ~\cite{wornow2023ehrshot}.
For MIMIC-IV, we construct patient-level splits; for EHRSHOT, we adopt the official splits and task definitions stated in the original work. All experiments are reproducible.

\subsection{EHR Data Representation Framework}\label{sec:rep_framework}
In our work, three representation frameworks are considered:

\textbf{Multivariate time-series.} A patient's record is transformed into a matrix $X \in \mathbb{R}^{T \times D}$, where $T$ is the number of discrete time steps and $D$ is the number of selected clinical features. This matrix is constructed by aligning raw clinical events to a 1-hour grid, with both forwards and backwards filling, followed by population-median imputation~\cite{wu2024instruction}. For MIMIC-IV, $T=24$; for EHRSHOT, we use a 7-day observation window prior to hospital discharge, hence $T=168$. Data from EHRSHOT cover multiple clinical scenarios, including primary care, the emergency department, hospital admissions, and ICU admissions~\cite{mesinovic2025dynagraph}. Samples are padded or truncated as necessary. Additionally, we exclude events with more than 90\% missing timestamps across the entire dataset by default. For feature selection experiments, we adjust the missing rate threshold as needed.

\textbf{Event stream.} Define the event stream of a patient as a sequence $P = (e_1, \dots, e_N)$, where each event is a tuple $e_i = (t_i, \text{code}_i, \text{value}_i)$. An event's code and its categorised value are used to determine a single token; this token's embedding ($E_{\text{Token},i}$) is then concatenated with a positional embedding ($E_{\text{POS},i}$) to form the final input vector $[E_{\text{Token},i} || E_{\text{POS},i}]$ for that event, as shown in Fig.~\ref{fig:rep}. 

\textbf{Textual event stream.} We transform event streams into natural language while preserving chronological order of events, as shown in Fig.~\ref{fig:rep}. Concurrent events are explicitly marked as occurring at the same time in the text. Event codes are assigned to human-readable descriptions based on clinical coding systems, with a specific descriptive template for each type of event that ensures consistency with existing work~\cite{zhu2025medtpe, fleming2024medalign, wu2024instruction}. Numerical values are appended with units to enhance clinical interpretability.


\section{Experiment Results}

\subsection{Experiment Setup}
To benchmark the structurally distinct EHR representations, we selected models tailored to each data format:
\emph{(A) Multivariate time-series matrix:} We evaluate Transformer~\cite{vaswani2017attention}, Multilayer Perceptron (MLP)~\cite{taud2017multilayer}, LSTM~\cite{yu2019review}, and RETAIN~\cite{choi2016retain}. Models were trained with a cross-entropy loss and used early stopping on validation metrics.
\emph{(B) Event stream:} We evaluate CLMBR~\cite{steinberg2021language, wornow2023ehrshot}, a foundation model pre-trained on 2.57M EHR data using the OMOP vocabulary, which is compatible with the EHRSHOT dataset. To ensure compatibility for MIMIC-IV, we implemented a multi-step pipeline with Athena OMOP vocabularies~\cite{reich2024ohdsi} to convert the medical code in MIMIC-IV into the required format, leveraging a series of open-source tools~\cite{ohdsi-mimic, meds-etl, femr-0-1-16, wornow2023ehrshot} with label extraction modules for \emph{ICU mortality} and \emph{ICU phenotyping} tasks as defined in Sec.~\ref{sec:task_def}. Another baseline is Count, an XGBoost based on event countst~\cite{chen2016xgboost}.
\emph{(C) Textual event stream:} This representation was evaluated using a suite of 8B to 20B LLMs, including GPT-OSS-20B~\cite{openai2025gptoss120bgptoss20bmodel}, Llama-3.1-8B-Instruct~\cite{llama3modelcard}, Qwen3-8B~\cite{qwen3technicalreport}, and DeepSeek-R1-as-Qwen3-8B~\cite{deepseekai2025deepseekr1incentivizingreasoningcapability}.

\begin{figure}[hb!] 
    \centering
    \includegraphics[width=1\linewidth]{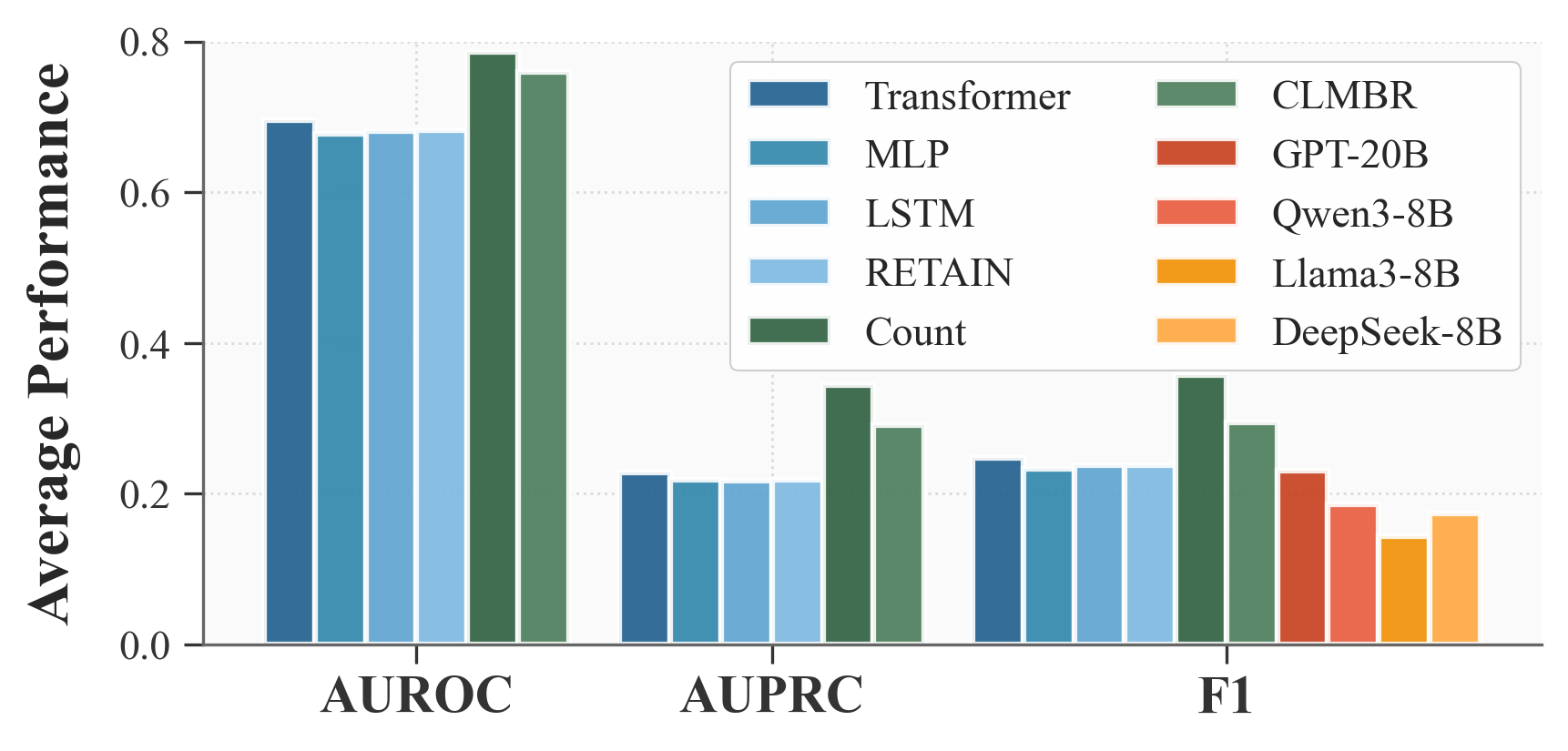}
    \caption{Avg. performance of models across ICU mortality, phenotyping, 30-day readmission, and 1-year pancreatic cancer predictions.}
    \label{fig:avg_performance}
\end{figure}

 For the few-shot training with the event stream representation, we set $k = 16$. For LLMs, the F1 score is used, which is considered as categorical rather than probabilistic. For binary tasks, we report AUROC, AUPRC, and F1 using a 0.5 threshold. For multi-label phenotyping, we report macro-AUROC, macro-AUPRC, and macro-F1.

\begin{figure*}[ht!] 
    \centering
    \includegraphics[width=1\textwidth]{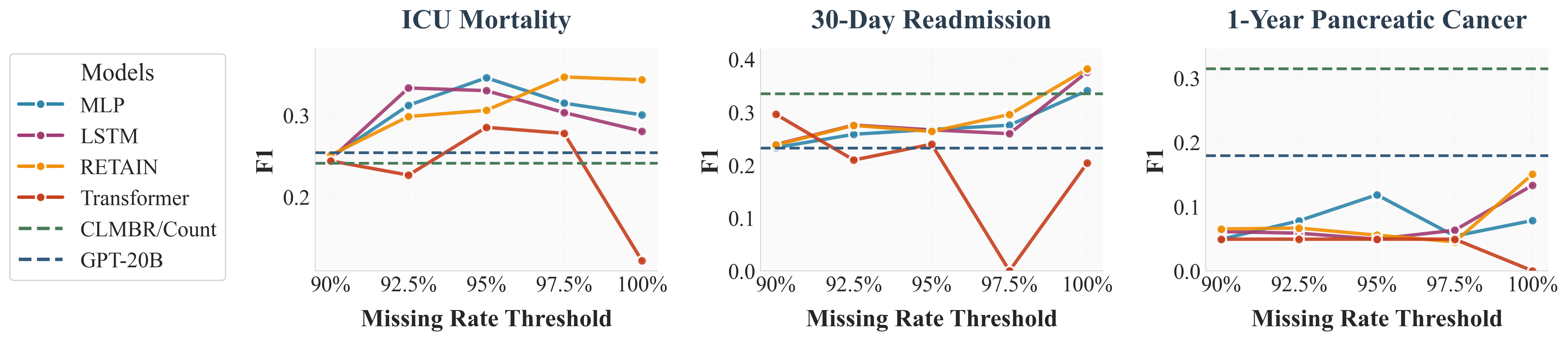}
    \caption{Comparison of model performance across different representations with different missing rate thresholds.}
    \label{fig:feature_selection}
\end{figure*}

\subsection{Comparison Across Representation Frameworks}

In this experiment, we systematically evaluate state-of-the-art (SOTA) methods across three EHR data representations with settings specified in Sec.~\ref{sec:rep_framework}.
As shown in Table~\ref{tab:main_results}, multivariate time-series models achieve moderate performance in both datasets. In MIMIC-IV ICU mortality, F1 scores range from 0.244 (Transformer) to 0.251 (RETAIN), and for ICU phenotyping from 0.387 (MLP) to 0.401 (Transformer). In EHRSHOT, performance is even lower: for 30-day readmission, F1 scores range from 0.233 (MLP) to 0.296 (Transformer), and for 1-year pancreatic cancer, they remain below 0.07 in all models. These results suggest that, while time-series models can capture temporal dynamics, they are constrained by high feature sparsity, irregular sampling, and dimensionality, particularly in the more complex longitudinal setting of EHRSHOT.

Although time-series models are the most widely adopted, event-stream models and LLMs with textual event streams consistently achieve the best average performance across tasks and datasets, as shown in Fig.~\ref{fig:avg_performance}. A potential reason is that time-series models rely heavily on task-specific feature selection~\cite{crone2010feature}, which is discussed further in Sec.~\ref{subsec:feature_selection}. Since we did not optimise feature selection for each prediction task, event stream models and LLMs that leverage all available features outperform the time-series models. This finding shows that event stream and textual event stream representations may serve as stronger general-purpose approaches for clinical prediction. 

In addition, LLMs with textual event streams are competitive in ICU mortality prediction. In the ICU scenario (MIMIC-IV), the best-performing LLM (GPT-20B) achieves a higher F1 score than all other approaches, demonstrating the potential of LLMs with textual event stream representations for short-term, high-frequency clinical scenarios. In the longitudinal care scenario (EHRSHOT), CLMBR-based models outperform LLMs. This reflects that structured event-stream models are better at capturing sparse but long-term relationships of EHR data~\cite{waxler2025generative}, while LLMs may struggle to understand the long and sparse context~\cite{ren2025comprehensive}. 

The effectiveness of pretrained models such as CLMBR is limited to the few-shot scenario. On MIMIC-IV, the Count model outperforms CLMBR in ICU phenotyping (F1 = 0.600 vs. 0.493). In contrast, CLMBR shows clear benefits in few-shot settings, as in EHRSHOT 30-day readmission (F1 = 0.335 vs. 0.266 for Count). These results suggest that a pretrained model is useful when data are scarce, but its advantage diminishes with sufficient training data.

\subsection{Impact of Feature Selection}
\label{subsec:feature_selection}

In this experiment, our objective is to evaluate how feature selection affects the performance of multivariate time-series models. Specifically, we apply feature selection based on the missing rate of features~\cite{chen2023dealing}, varying the threshold from 0.90 to 1.00 to progressively exclude features with higher missingness. We compare these models against the best-performing event stream model (CLMBR/Count) and the best textual event sequence model (GPT-OSS-20B), using the F1 score as the evaluation metric.
As shown in Fig.~\ref{fig:feature_selection}, moderate pruning helps multivariate time-series models on ICU tasks: F1 peaks when removing features with $>95–97.5$\% missingness, indicating that very sparse ICU variables add noise. Adjusting the missing rate threshold leads to large performance shifts, highlighting that the choice of features strongly influences the predictive accuracy of multivariate models.

An interesting observation emerges when comparing different clinical scenarios. In contrast to the ICU case, in the longitudinal care setting (EHRSHOT), the best performance is achieved when all features are retained, suggesting that sparse but informative features remain valuable in long-term patient trajectories. This difference can be explained by the nature of data collection in each setting. In ICU care, numerous redundant measurements are recorded due to patient severity, making careful feature selection beneficial~\cite{lee2015using}. In longitudinal care, however, clinicians typically request only essential measurements, making sparse features carry important information for clinical prediction~\cite{wornow2023ehrshot}.

\section{Conclusion}
\label{sec:conclusion}
This work provides the first benchmark across three EHR representation methods for clinical prediction tasks. Event-stream models consistently delivered the strongest results, with pretrained CLMBR excelling in few-shot settings and count-based models surpassing CLMBR in all-shot. Time-series models and LLMs are competitive on certain tasks (e.g., ICU mortality, 30-day readmission). The feature selection by missingness on time-series models offers favourable accuracy–complexity trade-offs on ICU tasks, while longitudinal outcomes benefit from retaining sparse but informative features; event-stream performance remains robust to these thresholds. The results of this benchmark offer practical guidance on selecting representations tailored to specific clinical scenarios. Future works include extending the benchmark to emergency and general inpatient settings, broadening the task suite, and calibrating model outputs to support uncertainty analysis across EHR representation methods.


\begingroup
\footnotesize
\setstretch{0.95}
\bibliography{refs}
\bibliographystyle{IEEEtranN}
\endgroup

\end{document}